\documentclass[letterpaper, 10 pt, conference]{ieeeconf}
\IEEEoverridecommandlockouts
\usepackage{graphicx} \graphicspath{{figures/}}
\usepackage{times,latexsym}
\usepackage{amsmath,amssymb}
\usepackage{algorithmic}
\usepackage[linesnumbered,ruled,vlined]{algorithm2e}
\usepackage{acronym}
\usepackage{balance}
\usepackage{xspace}
\usepackage{setspace}
\usepackage[skip=3pt,font=small]{subcaption}
\usepackage[skip=3pt,font=small]{caption}
\usepackage[dvipsnames,svgnames,x11names]{xcolor}
\usepackage{booktabs,tabularx,colortbl,multirow,array,makecell}
\makeatletter
\let\NAT@parse\undefined
\makeatother
\usepackage[pagebackref,breaklinks,colorlinks]{hyperref}
\usepackage[capitalise,nameinlink]{cleveref}
\usepackage{overpic,wrapfig}
\usepackage{cite}
\usepackage[misc]{ifsym}
\usepackage{threeparttable}
\usepackage{tablefootnote}

\def\model{\textbf{\textsc{COLA}}\xspace}
\def\modelf{\textbf{\textsc{COLA-F}}\xspace}
\def\modell{\textbf{\textsc{COLA-L}}\xspace}

\newcommand{\cola}{\includegraphics[height=2.8ex]{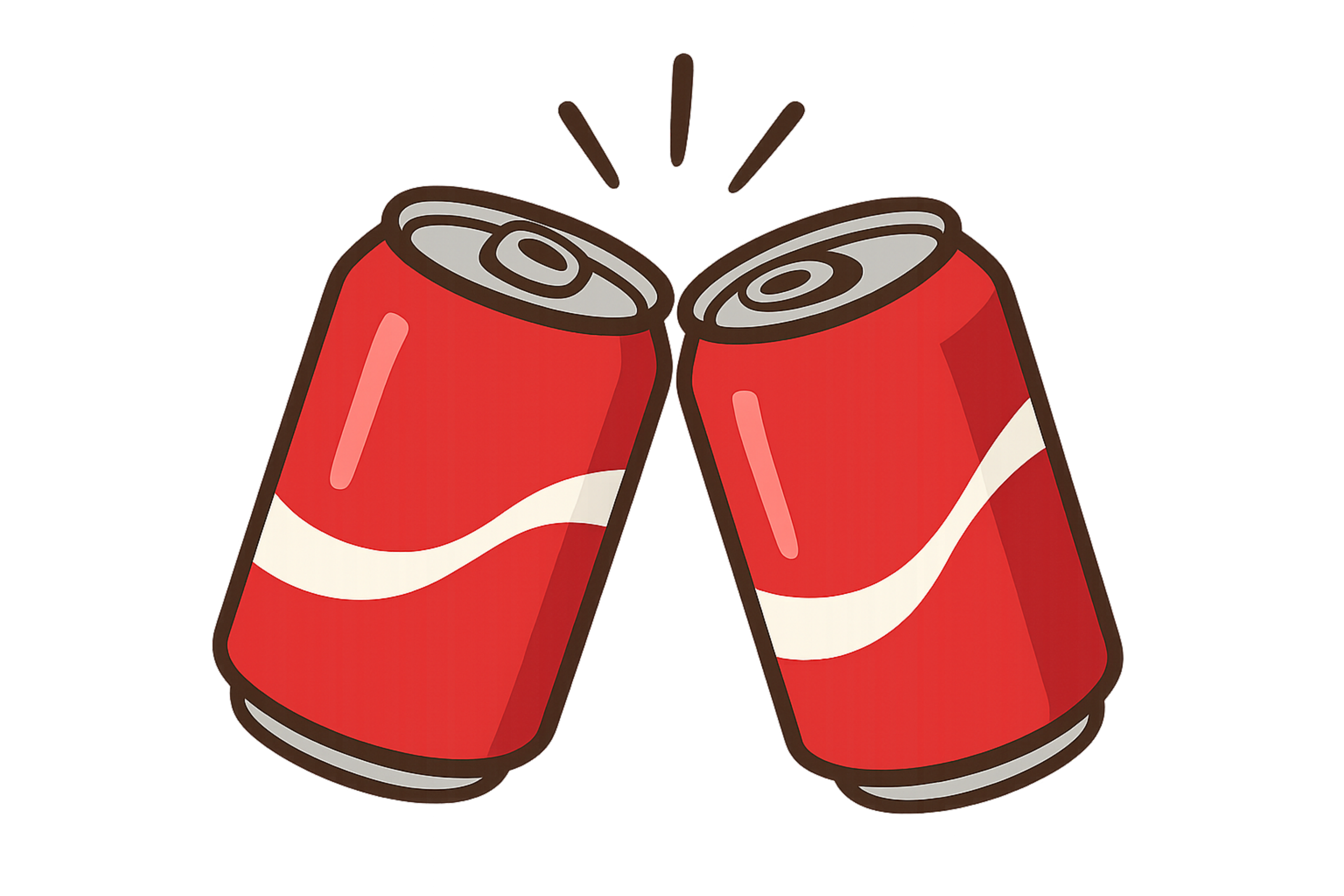}\hspace{2px}}

\usepackage{tabularx,booktabs,ragged2e}
\newcolumntype{J}{>{\RaggedRight\arraybackslash}X}

\definecolor{WeiColor}{rgb}{1,0.33,0.64}
\definecolor{myDarkBlue}{RGB}{55,81,139}
\definecolor{LimeGreen}{RGB}{158,186,217}
\definecolor{myLightGreen}{RGB}{126,171,85}
\definecolor{myRed}{RGB}{176,36,24}

\makeatletter
\DeclareRobustCommand\onedot{\futurelet\@let@token\@onedot}
\def\@onedot{\ifx\@let@token.\else.\null\fi\xspace}
\def\eg{\emph{e.g}\onedot}

\def\etc{\emph{etc}\onedot}

\makeatother

\crefname{algorithm}{Alg.}{Algs.}
\Crefname{algocf}{Alg.}{Algs.}
\crefname{section}{Sec.}{Secs.}
\Crefname{section}{Section}{Sections}
\crefname{table}{Tab.}{Tabs.}
\Crefname{table}{TABLE}{TABLES}
\crefname{figure}{Fig.}{Fig.}
\Crefname{figure}{Figure}{Figure}

\acrodef{llm}[LLM]{Large Language Model}
\acrodef{rl}[RL]{reinforcement learning}
\acrodef{mpc}[MPC]{Model Predictive Control}
\acrodef{mlp}[MLP]{Multi-Layer Perceptron}
\acrodef{mdp}[MDP]{Markov Decision Process}
\acrodef{ppo}[PPO]{Proximal Policy Optimization}
\acrodef{slerp}[SLERP]{Spherical Linear Interpolation}
\acrodef{wbc}[WBC]{Whole-Body Control}

\title{\LARGE \bf
\cola COLA: Learning Human-Humanoid Coordination for \\ Collaborative Object Carrying
}

\author{%
    \normalfont
    Yushi Du$^{\,^\ast1,2}$, Yixuan Li$^{\,^\ast2,3}$,  Baoxiong Jia\textsuperscript{\Letter}$^{\,^\ast2}$, Yutang Lin$^{\,2,4}$, Pei Zhou$^{\,1}$, \\
    Wei Liang\textsuperscript{\Letter}$^{\,3}$, Yanchao Yang\textsuperscript{\Letter}$^{\,1}$, Siyuan Huang\textsuperscript{\Letter}$^{\,2}$  \\
    \fontsize{8pt}{8pt}\selectfont $^1$ Department of Electrical and Electronic Engineering, the University of Hong Kong\\
    \fontsize{8pt}{8pt}\selectfont $^2$ State Key Laboratory of General Artificial Intelligence, BIGAI\\
    \fontsize{8pt}{8pt}\selectfont $^3$ School of Computer Science and Technology, Beijing Institute of Technology\\ 
    \fontsize{8pt}{8pt}\selectfont $^4$ Yuanpei College, Peking University\\
    \fontsize{8pt}{8pt}\selectfont $^\ast$ Denotes equal contributions \quad{} \fontsize{8pt}{8pt}\textsuperscript{\Letter} Corresponding authors 
}

\let\oldtwocolumn\twocolumn
\renewcommand\twocolumn[1][]{%
    \oldtwocolumn[{#1}{
    \begin{center}
            \includegraphics[width=\textwidth]{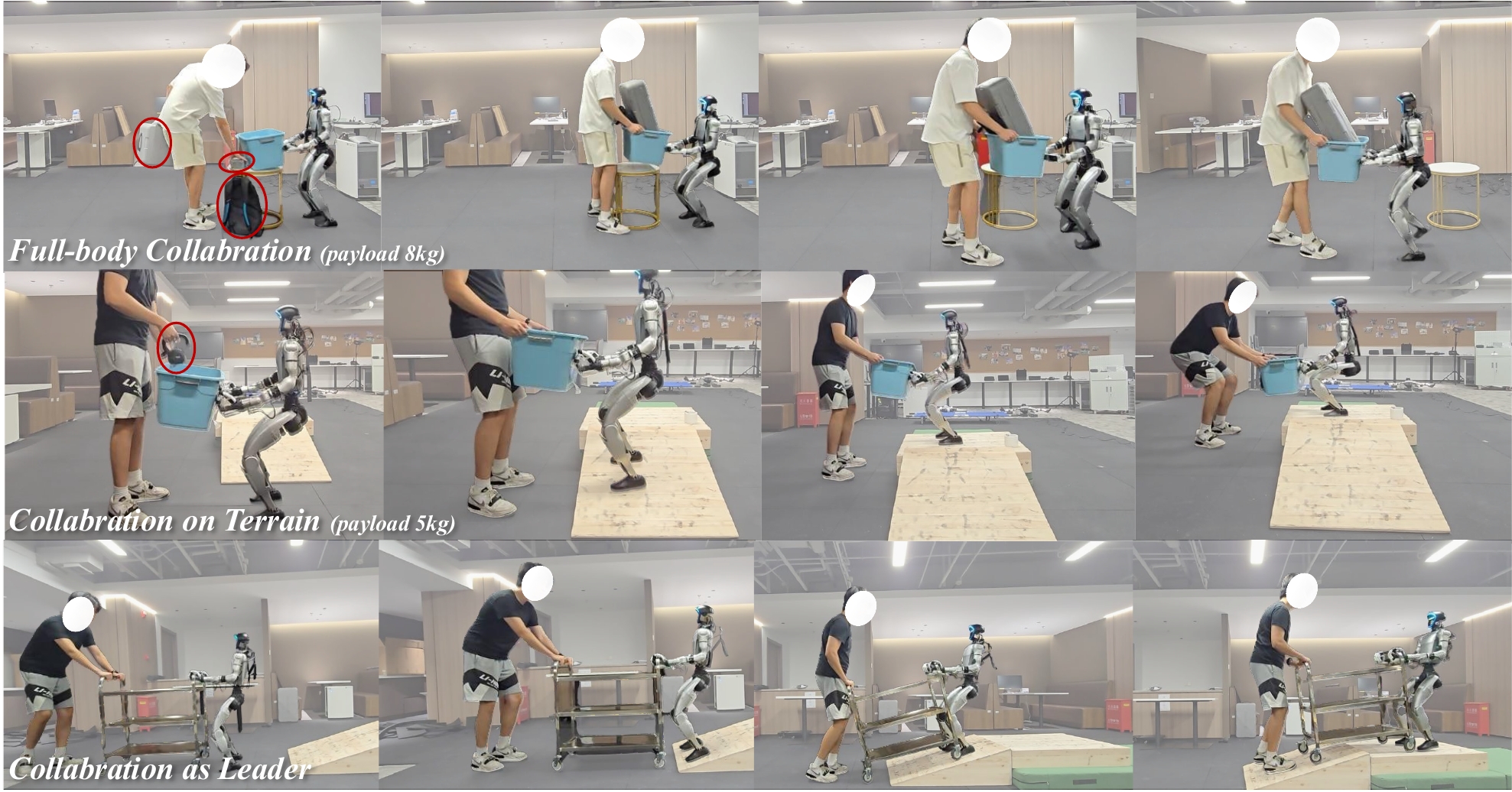}
            \captionsetup[figure]{hypcap=false}
            \captionof{figure}{\model provides a proprioception-only policy that enables compliant human-humanoid collaboration for carrying diverse objects across various movement patterns. (a) demonstrates horizontal trajectory coordination where the humanoid adapts to human motion patterns. (b) illustrates whole-body coordination during collaborative object-lowering tasks. (c) shows \model acting as the leader in collaboration with a human to drag a cart up a slope.}
            \label{fig:teaser}
        \end{center}
    }]
}

\begin{document}

\maketitle
\thispagestyle{empty}
\pagestyle{empty}
\bstctlcite{IEEEexample:BSTcontrol}

\begin{abstract}
% full-body, compliant； training setting for implicit learning， one policy two mode； terrain，save efforts
Human-humanoid collaboration shows significant promise for applications in healthcare, domestic assistance, and manufacturing. While compliant robot-human collaboration has been extensively developed for robotic arms, enabling compliant human-humanoid collaboration remains largely unexplored due to humanoids' complex whole-body dynamics.
In this paper, we propose a proprioception-only reinforcement learning approach, \model, that combines leader and follower behaviors within a single policy. The model is trained in a closed-loop environment with dynamic object interactions to predict object motion patterns and human intentions implicitly, enabling compliant collaboration to maintain load balance through coordinated trajectory planning.
We evaluate our approach through comprehensive simulator and real-world experiments on collaborative carrying tasks, demonstrating the effectiveness, generalization, and robustness of our model across various terrains and objects.
Simulation experiments demonstrate that our model reduces human effort by $24.7\%$. compared to baseline approaches while maintaining object stability. Real-world experiments validate robust collaborative carrying across different object types (boxes, desks, stretchers, \etc) and movement patterns (straight-line, turning, slope climbing). Human user studies with $23$ participants confirm an average improvement of $27.4\%$ compared to baseline models.
Our method enables compliant human-humanoid collaborative carrying without requiring external sensors or complex interaction models, offering a practical solution for real-world deployment. Our project website is available at \url{https://yushi-du.github.io/COLA/}.

\end{abstract}

\section{Introduction}
\label{sec:Introduction}
% 1. the importance of collaboration:
% With the rapid advancement of contemporary robotics technology, the demand for integrating robots into daily human life to assist in a variety of tasks is steadily increasing. Among the diverse tasks requiring human-humanoid collaboration, collaborative carrying stands out due to its frequent occurrence in both everyday life and industrial contexts, thus possessing substantial research significance. Humanoid robots, owing to their object manipulation and transport capabilities that closely resemble those of humans, as well as their similar responses to physical laws, are particularly well-suited for fulfilling the requirements of this task.
% With the rapid advancement of humanoid robots, the demand for enabling robots to perform versatile tasks is steadily increasing. While humanoids can excel at various locomotion~\cite{ma2025styleloco}, teleoperation~\cite{li2025clone,ze2025twist}, and manipulation tasks, human-humanoid collaboration capabilities still need significant improvement.
% Although humans can easily infer object dynamics and partners' intent through subtle object movement cues, adapting their actions to maintain stable load sharing and reduce physical effort, humanoid robots' ability to perform such compliant collaboration remains largely unexplored.
Recent years have witnessed significant progress in humanoid robot development, including agile locomotion~\cite{ma2025styleloco, zhuang2024parkour, gu2025humanoid, zhang2025hub}, teleoperation~\cite{li2025clone,ze2025twist}, and dexterous manipulation\cite{ze2025generalizable,shah2025mimicdroid}.
While these advances highlight the growing versatility and robustness of humanoid control, progress in enabling humanoid robots to collaborate effectively with humans remains limited. Human-robot collaborations have been a long-standing challenge~\cite{rozo2016learning, rahem2022human, yu2021adaptive, zhu2024you, elwin2022human, zhi2025closed}, requiring the modeling of diverse human behaviors, adaptive responses to dynamic interactions, and coordinated planning for shared tasks. As humanoid robots develop increasingly reliable motor and control abilities, we argue that addressing collaboration is both timely and essential to realizing their central role in supporting human life.

Object carrying~\cite{agravante2019human,bethala2025collaboration,gonzalez2025nonlinear} has become a representative task for advancing human-robot collaboration. Its core challenges arise from adapting to diverse environments (\eg, maintaining stable support of objects across varying terrains), responding compliantly to human motions (\eg, standing up together) often with limited or no direct force sensing, and dynamically allocating roles such as leading or following to improve efficiency. These interdependent requirements make the task particularly difficult for humanoids, as true collaboration requires integrating all aspects to ease the human partner, rather than addressing a single constraint as in prior work on environment-conditioned locomotion~\cite{ma2025styleloco, zhuang2024parkour, zhang2025falcon}, compliance behavior learning~\cite{wong2023vision,zhi2025learning, yu2021adaptive, xu2025facet}, or high-level intention prediction in open-loop object-finding or serving tasks~\cite{gao2021hybrid,sheng2025human,liu2025idagc,zhi2025closed}. Addressing these challenges requires a policy that unifies force interactions, implicit constraints, and dynamic coordination into a coherent framework for humanoid collaborative carrying.

% Recently, several researchers have investigated compliant robot-human collaboration~\cite{gao2021hybrid,wong2023vision,gonzalez2025nonlinear,agravante2019human} based on robotic arms, advancing the state of the art in this field.
% However, current works for robot-human collaboration using humanoid robots typically focus on following predicted human intentions or responding to applied external forces, training policies in open-loop environments without carrying objects. This approach ignores compliance requirements during the collaboration process and the complex dynamics introduced by shared object manipulation.
% Moreover, these methods often adopt model-based approaches to perform predefined skills or commands, limiting the generalization of collaboration across various interaction patterns and terrain variations while neglecting full-body coordination capabilities.

To address these challenges, we propose a learning-based policy for human-humanoid collaborative carrying that leverages reinforcement learning to model dynamic and versatile interactions. The policy allows humanoid robots to share loads with humans in a compliant manner while flexibly switching roles between leader and follower. Our design is built on two key insights: (i) offsets between joint states and their targets provide a proxy for estimating interaction forces, and (ii) the carried object's state encodes implicit collaboration constraints such as stability and coordination. To incorporate these, we adopt a teacher-student framework in which the teacher policy, trained with both proprioceptive and privileged object-state information, is learned with rewards on the humanoid motion (\eg, robust locomotion across varied terrains) and the object status (\eg, maintaining a stretcher). The student policy, distilled from the teacher, relies solely on proprioceptive inputs for real-world inference and deployment. Role allocation is controlled via a velocity command, where zero velocity corresponds to following. 
% In this paper, we aim to develop \textbf{an end-to-end policy for humanoid–human collaborative carrying that enables humanoids to share loads with humans in a compliant manner}, without relying on additional sensors. We propose a three-stage residual learning framework combined with a closed-loop training environment that explicitly models object dynamics.
% We first train a base locomotion policy that enables the humanoid t   o perform robust whole-body movement. Then we train a residual teacher policy on top of the pre-trained locomotion policy with privileged access to object state information, which facilitates learning of collaborative behaviors. This knowledge is then distilled into a unified student policy that operates solely on proprioception, implicitly adapting to interaction forces and coordinating the humanoid's full-body motion with human partners. The policy supports both leader and follower modes, enabling flexible object transport through compliant collaboration.

We conduct extensive simulation experiments to demonstrate that \model reduces human effort by $31.47\%$ in collaborative carrying tasks compared to baseline approaches. Trajectory analysis shows that our method achieves $10.2$ cm/s mean linear velocity tracking error and $0.1$ rad/s mean angular tracking error relative to human motion, indicating precise coordination.
Real-world experiments validate that \model successfully tracks human movement patterns while assisting with object lifting, lowering, and transport along both straight and curved trajectories. The intention of human are implicitly learned by simple pushing and pulling actions, without additional commands from remote controls.
Human user studies with $23$ participants \textbf{confirm compliant collaborative carrying across diverse scenarios, demonstrating the practical effectiveness of our approach for human-robot collaborative object transport}.

Overall, our contributions can be summarized as follows:
\begin{itemize}
    % \item We propose a vision-free two-stage residual learning framework for whole-body human-humanoid collaborative carrying that uses only proprioceptive feedback to enable coordinated object transport across diverse terrains and movement patterns.
    % \item We introduce a novel proprioception-based policy that learns to predict human intentions and adapt humanoid motion without external sensors or explicit human motion models.
    % \item We enable humanoid robots to perform complex collaborative carrying tasks (lifting, lowering, curved trajectories)
    % \item We propose a unified model that focuses on solving the human-humanoid collaborative carrying task, extending the capability range of humanoid robots to broader domains so that they can effectively assist humans in practical tasks. For instance, to carry objects that are oversized or have complex shapes, and demonstrate proficiency in various motion modes, including lifting, lowering, and following curved trajectories.
    % \item We develop a reinforcement learning-based training framework that enables the robot to infer human actions from the motion state information of the object, and assists humans effectively in collaborative carrying tasks.
    \item We propose a unified residual model that relies solely on proprioception for whole-body collaborative carrying, enabling compliant, coordinated, and generalizable collaboration across diverse movement patterns.
    \item We develop a three-step training framework and closed-loop training environment that explicitly models humanoid-object interactions, enabling the robot to implicitly learn object movements and assist humans through compliant collaboration.
    \item We demonstrate our policy in simulation and real-world settings, where our method achieves superior effort reduction and trajectory coordination compared to baseline approaches. The human user study confirms that our model achieves more compliant collaboration.
\end{itemize}

\begin{figure*}[ht!]
    \centering
    \includegraphics[trim= 6.7cm 5.2cm 7cm 5cm, clip, width=\linewidth]{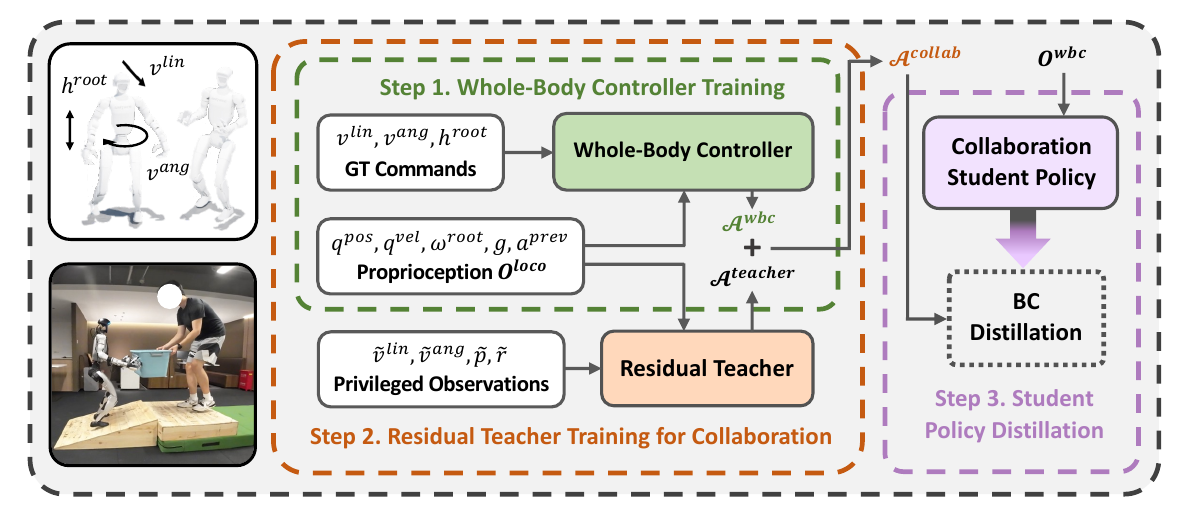}
    % \caption{\textbf{Overview.} Our policy is trained using a teacher-student scheme. The teacher policy is learned with privileged observation and proprioception, guiding the learning process of the student policy. The actions are executed by a PD controller.}
    \vspace{-20pt}
    \caption{\textbf{Overview of \model.} Our Policy mainly consists of three steps: (i) We train a base whole-body control policy to provide a robust whole-body controller. (ii) In the closed-loop training environment, we train a residual teacher policy on top of the whole-body control policy with privileged information for human-humanoid collaboration. (iii) We distill the knowledge from the teacher policy into a student policy for real-world deployment using behavioral cloning. }
    \label{fig:pipeline}
    \vspace{-15pt}
\end{figure*}

% 2. brief summarize of previous humanoid works to collaboration:
% Previous works in extending the capabilities of humanoid to interact with human 

% 3. limitations of previous works and our contributions:

% 4. three points:

\section{Related Work}
\label{sec:Related Work}
\subsection{Robot-human Collaboration}
Robot-human collaboration has been a long-standing research topic spanning from robotic arms to legged robots~\cite{li2025human,sheng2025human,liu2025idagc}. While robotic arm-based collaboration systems typically assist humans in confined workspaces, researchers increasingly employ humanoid robots to provide assistance in open environments, leveraging their superior mobility and human-like morphology.
However, current human-humanoid collaboration methods~\cite{rahem2022human,zhu2024you,elwin2022human} rely primarily on model-based approaches. Existing works~\cite{agravante2019human,agravante2014collaborative} use heuristic rules that predefine a set of subtasks, including basic walking patterns, and identify primitive behaviors necessary for collaborative carrying. Some works~\cite{palinko2016robot,nicolis2018human} focus on predicting human intention from multi-modal data and performing subtasks accordingly. While H$2$-COMPACT~\cite{bethala2025collaboration} proposes a learning-based model that uses haptic cues to predict horizontal velocity commands, it still operates with limited scope.
All these approaches neglect whole-body coordination capabilities in human-humanoid collaborative carrying~\cite{wang2018facilitating}, thus lacking the ability to perform complex collaborative tasks such as picking up objects from the ground or carrying objects while climbing slopes.
In this work, we propose a residual learning framework that enables humanoids to collaborate with humans using whole-body coordination, significantly broadening the range of collaborative carrying scenarios that humanoid robots can handle.

\subsection{Compliant Whole-body Control}
Position-only control lacks the compliance required for human-humanoid interaction~\cite{yu2021adaptive,hartmann2024deep}, as it operates without force awareness. Force regulation is crucial for collaborative tasks~\cite{li2025impedance}, particularly those involving human contact.
Recent research~\cite{zhi2025learning} has demonstrated the effectiveness of force and compliance control in contact-rich manipulation tasks~\cite{du2023learning}. These methods explicitly estimate contact forces and integrate them into control policies, achieving improved performance on tasks such as force tracking and compliant responses to varied force and position inputs. Moreover, other approaches~\cite{xu2025facet} learn force characteristics implicitly to enable compliant and force-adaptive behaviors on legged robots.
While these advances demonstrate that incorporating force feedback provides significant advantages for robotic interaction tasks, how force-aware control benefits human-humanoid collaboration remains underexplored. Building on these insights, we implicitly incorporate force considerations into our human-humanoid collaborative framework, enabling more natural and intuitive cooperative interactions.

\section{Methodology}
\label{sec:Methodology}
\subsection{Overview}
% We present the task of human-humanoid Collaborative Carrying, which can be formulated as requiring the agent to assist in carrying an object of a specific shape or size that is difficult for a single person to transport. Given that a human operator is already carrying the object, the agent’s objective is to minimize the velocity difference between itself and the human operator while maintaining the object in the desired pose (\eg, keeping a box or stretcher horizontal to the ground plane). Due to the practical demand for transporting large objects in everyday life, we believe this task holds significant research value.
We define the task of human-humanoid Collaborative Carrying as a humanoid assisting a human partner to transport an object that is challenging for a single person due to its size or weight.
We assume the human partner is engaged in carrying the object, and the robot's objectives are three-fold: (i) to coordinate its movement by aligning with the human's velocity, (ii) to support the object's weight, thereby reducing the human's physical burden, and (iii) to stabilize the object's orientation throughout the transportation.

% Our training pipeline consists of a three-step reinforcement learning stage, which is composed of \textbf{a. Vanilla locomotion policy}, \textbf{b. Residual teacher policy for collaboration} and \textbf{c. Distillation}.
% Our training pipeline consists of three reinforcement learning stages: \textbf{a. Vanilla locomotion policy}, \textbf{b. Residual teacher policy for collaboration,} and \textbf{c. Knowledge distillation to a student policy}.
% The following section details this three-stage design.
Our training pipeline is composed of three distinct learning steps: \textbf{a. Whole-body controller training}, \textbf{b. Residual teacher policy training for collaboration}, and \textbf{c. Student policy distillation}. We elaborate on this three-step design in the subsequent section.

\subsection{Whole-body Control Policy}
In the first step, we train a \ac{wbc} policy with no additional constraints in the simulator, where the goal command is $\mathcal{G} = [\mathcal{G}^{\text{lower}}, \mathcal{G}^{\text{upper}}]$, which consists of a lower-body locomotion goal command $\mathcal{G}^{\text{lower}}_t \triangleq [v^{\text{lin}}_{t}, v^{\text{ang}}_{t}, h^{ \text{root}}_{t}]$ with velocity and height commands and an upper-body end-effector goal command $\mathcal{G}^{\text{upper}}_t = [p^{\text{ee}}, r^{\text{ee}}]$ specifying target position and rotation. The observations of the \ac{wbc} policy are defined as $\mathcal{O}^{\text{wbc}}_{t} \triangleq [q^{\text{pos}}_{t-l:t}, q^{\text{vel}}_{t-l:t}, \omega^{\text{root}}_{t-l:t}, g_{t-l:t}, a^{\text{prev}}_{t-(l+1):t-1}]$ where $l$ is the length of history observation, $q^{\text{pos}}_{t} \in \mathbb{R}^{N}$ denotes joint positions, $q^{\text{vel}}_{t} \in \mathbb{R}^{N}$ denotes joint velocities, $\omega^{\text{root}}_{t} \in \mathbb{R}^{4}$ denotes robot root orientation, $g_{t} \in \mathbb{R}^{3}$ for gravity in the robot root frame and $a^{\text{prev}}_{t} \in \mathbb{R}^{n}$ denotes previous actions under the corresponding time frame. The action space, $\mathcal{A}^{\text{wbc}}$, represents the target joint positions of the $N=29$ joints of the G1 robot, excluding the fingers. We use PD position control for actuation.

A robust \ac{wbc} policy serves as the foundation for the complex task of human-humanoid collaborative carrying.
Building upon previous studies~\cite{zhuang2024parkour,zhang2025falcon}, which have demonstrated the efficacy of reinforcement learning for training humanoid robot loco-manipulation and have achieved notable success in real-world deployments, we adopt the \ac{ppo} algorithm to conduct reinforcement learning-based training for humanoid collaboration in our research. 
Specifically, we train the \ac{wbc} policy taking both lower-body locomotion commands $[v^{\text{lin}}, v^{\text{ang}}, h^{\text{root}}]$ and upper-body end-effector commands $[p^{\text{ee}}, r^{\text{ee}}]$. 
The end-effector pose commands, in particular, are generated from a sampling space detailed in \cref{sec:details}. The policy is formally defined as:
$$\mathcal{F}^{\text{wbc}}: \mathcal{G} \times \mathcal{O}^{\text{wbc}} \rightarrow \mathcal{A}^{\text{wbc}},\: \mathcal{A}^{\text{wbc}} \in \mathbb{R}^N.$$
We train the \ac{wbc} policy with rewards following prior works~\cite{LeggedLab,zhang2025falcon}.
To improve the robustness under payloads, we apply external forces to the humanoid's end-effectors during training, improving its force-adaptive capabilities. 

% \subsection{Collaboration Policy}
\subsection{Residual Teacher Policy}
\begin{figure}[t!]
    \centering
    \includegraphics[width=\linewidth]{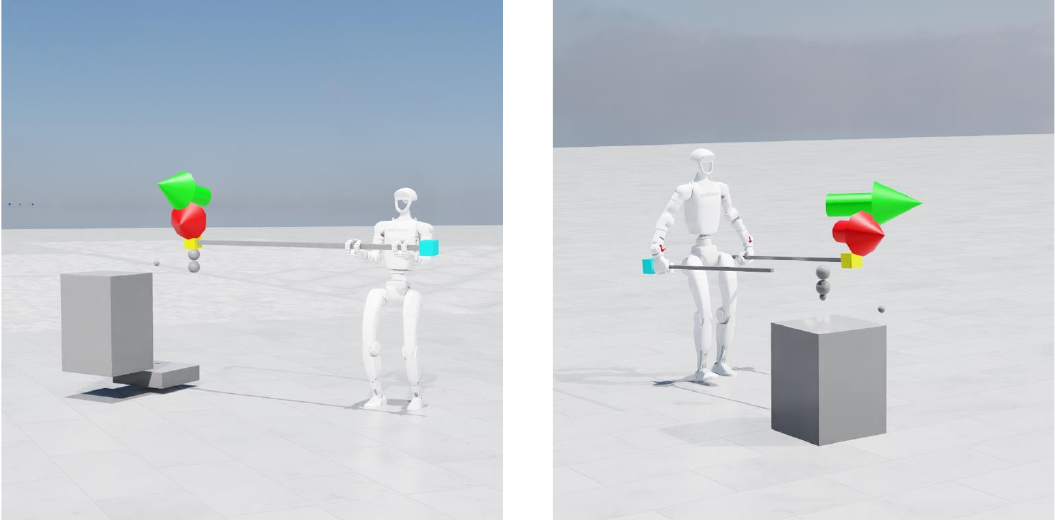}
    \caption{\textbf{Closed-loop Training Environment}. This figure illustrates our closed-loop training environment in simulation. The green arrow represents the goal velocity of the carried object, while the red arrow indicates its current velocity.}
    \vspace{-15pt}
    \label{fig:training_setting}
\end{figure}

In the second training step, we introduce a closed-loop environment for policy training to explicitly model the dynamic interaction between the human, object, and humanoid.
For instance, by placing a simulated box between the robot’s hands, the hands are forced to move in a particular manner, as illustrated in \cref{fig:training_setting}. 
Under these constraints, we train a residual teacher policy on top of the pre-trained \ac{wbc} policy in the closed-loop training environment. 
The residual teacher policy takes observations $\mathcal{O}^{\text{teacher}}_{t} \triangleq [\mathcal{O}^{\text{wbc}}_{t}, \mathcal{O}^{\text{priv}}_{t}]$ as inputs, where the privileged component $\mathcal{O}^{\text{priv}}_{t}$ consists of $\mathcal{O}^{\text{priv}}_{t} \triangleq [\widetilde{v}^{\text{lin}}_{t-l:t}, \widetilde{v}^{\text{ang}}_{t-l:t}, \widetilde{p}_{t-l:t}, \widetilde{r}_{t-l:t}]$ represent the GT states of the carried object with a history of length $l$.

Effective human-humanoid collaboration hinges on the robot's ability to adapt its collaboration strategy in real-time to the object's changing dynamics.
To this end, we formulate the collaboration policy as a residual policy atop the base whole-body control policy, enabling the model to implicitly infer collaborative intent directly from physical interaction. 
We hypothesize that this implicit learning paradigm is highly effective for two primary reasons. First, the exploration space for object dynamics during whole-body controller training may not fully encompass the motion constraints present in real-world collaboration. Second, the nuances of physical interaction are difficult to encode with manually designed, explicit commands. We validate the efficacy of our approach through ablation studies detailed in \cref{sec:experiments}.
% To evaluate the efficiency of this implicit learning paradigm, we conduct ablation experiment to be discussed in \cref{sec:experiments}. We attribute this efficiency to two main factors: first, the exploration space for object dynamics during locomotion may not fully encompass the motion constraints present in real-world collaboration; second, manually designed explicit commands may fail to adequately represent the required instructions. Detailed descriptions of these experiments are provided in \cref{sec:experiments}. 

The second step of our pipeline, illustrated in \cref{fig:pipeline}, involves training a residual teacher policy to build collaborative skills upon the base whole-body controller. 
To accurately model the object's dynamics, this teacher is granted access to privileged information ($\mathcal{O}^{\text{priv}}$), which includes the object's ground-truth pose and velocity history.
% [\widetilde{v}^{\text{lin}}_{t-l:t}, \widetilde{v}^{\text{ang}}_{t-l:t}, \widetilde{p}_{t-l:t}, \widetilde{r}_{t-l:t}]
The teacher policy, $\mathcal{F}^{\text{teacher}}$, leverages both privileged information and the robot's proprioception ($\mathcal{O}^{\text{wbc}}$) to output a \textit{residual action}, $\mathcal{A}^{\text{teacher}}$. This action represents a corrective adjustment to the whole-body controller's output. The policy is defined as:
$$\mathcal{F}^{\text{teacher}}: [\mathcal{O}^{\text{wbc}}, \mathcal{O}^{\text{priv}}] \rightarrow \mathcal{A}^{\text{teacher}},\: \mathcal{A}^{\text{teacher}} \in \mathbb{R}^N,$$
$$\mathcal{A}^{\text{collab}} = \mathcal{A}^{\text{wbc}} + \mathcal{A}^{\text{teacher}}.$$ 
% To enable the policy to represent the dynamics of the transported object more accurately, we provide the object's state including its pose and velocity $\widetilde{v}^{lin}_{t-l:t}, \widetilde{v}^{ang}_{t-l:t}, \widetilde{p}_{t-l:t}, \widetilde{r}_{t-l:t}$ as privileged observation inputs to the teacher policy. 
% These observations encapsulate key target information relevant to the collaboration process, enabling the model to learn the target speed, height, and other objectives that the object should achieve under the current state. The action space of $\mathcal{A}^{teacher}$ is the same as $\mathcal{A}^{loco}$, which is the target joint position of the robot.
% $$\mathcal{F}^{teacher}: [\mathcal{O}^{loco}, \mathcal{O}^{priv}] \rightarrow \mathcal{A}^{teacher},\: \mathcal{A}^{teacher} \in \mathbb{R}^n,$$
% $$\mathcal{A}^{collab} = \mathcal{A}^{loco} + \mathcal{A}^{teacher}.$$ 

During this step, the goal command of the model is modified based on the settings described in \cref{sec:details}. The teacher's learning is guided by a composite reward function that combines base whole-body control rewards with task-specific rewards, as detailed in \cref{tab:reward}.

\subsection{Knowledge Distillation}
In the distillation step, we transfer the expertise of the teacher policy ($\mathcal{F}^{\text{wbc}} + \mathcal{F}^{\text{teacher}}$) into a student policy, $\mathcal{F}^{\text{student}}$, designed for real-world deployment. This student policy operates without access to privileged information, relying solely on the proprioceptive observations, $\mathcal{O}^{\text{wbc}}$:
\[
\mathcal{F}^{\text{student}}: \mathcal{O}^{\text{wbc}} \rightarrow \mathcal{A}^{\text{student}},\: \text{where } \mathcal{A}^{\text{student}} \in \mathbb{R}^N.
\]
We use behavioral cloning to distill the teacher policy into a student policy, training the student to mimic the teacher by minimizing the mean squared error between their outputs during interactions with the environment. The optimization is defined by the following loss function:
\[
\mathcal{L}_{\text{distill}} = \mathbb{E} \left[ \Vert \mathcal{A}^{\text{student}} - \mathcal{A}^{\text{collab}}\Vert^2 \right].
\]

% \subsection{Leader and Follower Setting}
Furthermore, we define two distinct experimental settings based on whether the model observes the goal command during collaboration: \modelf \textbf{(Follower)} and \modell \textbf{(Leader)}. In the \modelf setting, all networks receive a goal command input of zero, whereas in the \modell setting, the policy is provided with a sampled goal command within the same range used for the whole-body controller.

\begin{table}[t!]
    \raggedright
    \caption{\textbf{Reward Functions} for collaboration policy training. }
    \label{tab:reward}
    \begin{threeparttable}
        \begin{tabularx}{\linewidth}{Jcc}
            \toprule
            Term  & Expression & Weight \\
            \midrule
            % \ac{wbc} Rewards & $\mathcal{R}^{\text{wbc}}$ & $34.77$ \\
            Linear Vel. Tracking & $\phi(v_{\text{lin}}^{\text{CoM}} - v_{\text{lin}}^{\text{applied}})$ & $1.0$\\
            Yaw Vel. Tracking & $\phi(v_{\text{ang}}^{\text{CoM}} - v_{\text{ang}}^{\text{goal}})$ & $1.0$\\
            Z-axis Vel. Penalty & $-\| v^{\text{obj}}_z \|$ & $0.05$\\
            Height Diff. Penalty & $-\| h^{\text{obj}}_1 - h^{\text{obj}}_2 \|$ & $10.0$\\
            Force Penalty & $-|\mathcal{F}^{\text{support-obj}}|$ & $0.002$ \\
            \bottomrule
        \end{tabularx}
    \begin{tablenotes}[flushleft]
      \footnotesize
      \item*Note: $v^{\text{applied}}_{\text{lin}}$ is the applied end-point linear-velocity(\cref{sec:details}); $v^{\text{goal}}_{\text{ang}}$ is the target angular velocity of the object; $h^{\text{obj}}_1$, $h^{\text{obj}}_2$ are the predefined heights of the object's far ends; $\mathcal{F}^{\text{support-object}}$ is the horizontal force between the support body and object; $\phi(x) = e^{-\|x\|}$. 
    \end{tablenotes}
    \end{threeparttable}
    \vspace{-2pt}
\end{table}

\begin{table}[t!]
    \raggedright
    \caption{\textbf{Command Sampling Ranges.} 
    The sampling range of our command. This table mainly consists of whole-body control commands, collaborative carrying commands.
    %This table illustrates the sampling range of our goal command. The terms that start with Locomotion stand for the command for the locomotion environment, while the terms that start with Support Object stand for the command for the collaboration environment. The EE Postion denotes the length of the box where the end-effector position goal sampled from. “The EE Orientation parameter specifies the half-angle of the cone that defines the sampling range of the end-effector orientation goal.
    }
    \label{tab:command}
        \begin{threeparttable}
        \begin{tabularx}{\linewidth}{Jc}
            \toprule
            Term  & Range \\
            \midrule
            Base Lin. Vel. X $(m/s)$ & $(-0.8, 1.2)$ \\
            Base Lin. Vel. Y $(m/s)$ & $(-0.5, 0.5)$ \\
            Base Ang. Vel. $(rad/s)$ & $(-1.2, 1.2)$ \\
            Base Height $(m)$ & $(0.45, 0.9)$ \\
            End-effector Position $(m)$ & $(0.15)$ \\
            End-effector Orientation $(rad)$ & $(\pi/6)$ \\
            Support Object Lin. Vel. $(m/s)$ & $(-0.6, 1.0)$ \\
            Support Object Ang. Vel. $(rad/s)$ & $(-0.8, 0.8)$ \\
            Support Object Height $(m)$ & $(0.5, 0.85)$ \\
            \bottomrule
        \end{tabularx}
        \begin{tablenotes}[flushleft]
            \footnotesize
            \item*Note: End-effector Position denotes the side length of the cube where the goal position is sampled from; End-effector Orientation denotes the half-angle of the cone that defines the sampling range of orientation goals.
        \end{tablenotes}
        \end{threeparttable}
    \vspace{-15pt}
\end{table}

\section{Implementation Details}
\label{sec:details}
\subsection{Training Setup}
We train our policy in Isaac Lab on a single RTX $4090$D GPU using \ac{ppo} with $4096$ parallel environments. The actor and critic networks for the base \ac{wbc} policy are three-layer \acp{mlp} of size ($512$, $256$, $128$), while the residual teacher and studnet policy network employs two additional \acp{mlp} with the same dimensions. The training of the \ac{wbc}, residual teacher, and distillation policies takes $350$k, $250$k, and $250$k environment steps, respectively, which correspond to abot $15$k, $10$k, and $10$k \ac{ppo} update steps. The full training time is 48 hours.
% $3.36 \times 10^{5}$, $2.40 \times 10^{5}$, and $2.40 \times 10^{5}$ 
% environment steps for \ac{wbc}, residual teacher, and distillation, respectively. This corresponds to $1.40 \times 10^{4}$, $1.00 \times 10^{4}$, and $1.00 \times 10^{4}$ \ac{ppo} update steps respectively, with total training time of 48 hours.

\subsection{Observation Space Details}

We sample whole-body control commands from a pre-defined range. The end-effector goal command, representing the $6-$DoF target pose of the robot's wrist, is generated using \ac{slerp}. Since our task focuses on human-humanoid collaborative carrying rather than complex upper-body manipulation, the robot primarily needs to execute fine-scale arm adjustments to modify the object's pose and velocity. Therefore, we do not sample large-range upper-body motions. Instead, we sample end-effector goal commands in the vicinity of the default grasping pose, with positions randomly sampled within a small cubic region and orientations within a conical region around the nominal grasp orientation. We provide the sampling range details in \cref{tab:command}. Our whole-body controller achieves and tracking error of $5.6cm$ for end-effector goal position and $7^\circ$ for end-effector goal orientation.

\begin{table*}[t!]
    \centering
    \caption{\textbf{Quantitative evaluation in simulation.} We report results on velocity and height tracking, as well as the average external force between the robot and the carried object, to evaluate the effort required for joint carrying and movement.}
    \label{tab:main_exp_results}
    \resizebox{\linewidth}{!}{
        \begin{tabular}{ccccc}
            \toprule
            Methods  & Lin. Vel. ($m/s$) $\downarrow$ & Ang. Vel. ($rad/s$) $\downarrow$ & Height Err. ($m$) $\downarrow$ & Avg. E.F. ($N$) $\downarrow$  \\
            \midrule
            Explicit Goal Estimation & $0.235$ & $0.335$ & $0.102$ & $19.067$ \\
            \midrule
            % lin\_vel\_z & $-\|v_{b,z} \|^{2}$ & $-1.0$\\
            Transformer & $0.178$ & $0.310$ & $0.077$ & $19.382$ \\
            \modelf-History$10$ & $0.121$ & $0.131$ & $0.037$ & $15.435$ \\
            \modelf-History$50$ & $0.116$ & $0.132$ & $0.036 $& $14.574$ \\
            \modelf & $0.109$ & $0.118$ & \textbf{$0.031$} & $14.576$  \\
            \modell-History$10$ & $0.118$ & $0.106$ & $0.039$ & $13.924$ \\
            \modell-History$50$ & $0.112$ & $0.103$ & $0.036$ & $13.495$ \\
            \modell & \textbf{$0.102$} & \textbf{$0.098$} & $0.038$ & \textbf{$12.298$}  \\
            \bottomrule
        \end{tabular}
    }
\end{table*}

% \begin{table}[h]
% \centering
% \caption{List of Observations}
% \small
% \label{tab:observations}
% \begin{tabular}{@{}llll@{}}
% \toprule
% \textbf{Observation} & Notation & \textbf{Available To} & \textbf{Description} \\
% \midrule
% Command & $\mathbf{u}$ & Student \& Teacher & Includes $\qpos_{des}$, $K_p$, $K_d$, and virtual mass $\mathbf{m}$ \\
% Joint Positions & $\mathbf{q}$ & Student \& Teacher & Raw joint angles \\
% Joint Velocities & $\dot{\mathbf{q}}$ & Student \& Teacher & Angular velocities of joints \\
% Base Angular Velocity & $\omega$ & Student \& Teacher & Angular velocity of the base (from IMU) \\
% Projected Gravity Vector & $g$ & Student \& Teacher & Gravity vector projected into base frame \\
% \midrule
% Base Linear Velocity & $\mathbf{x}$ & Teacher Only & Linear velocity of the robot’s base \\
% Feet Contact States & $\mathbf{c}$ & Teacher Only & Binary indicators of foot-ground contact \\
% External Forces/Torques & $f_{\rm ext}$ & Teacher Only & Forces and torques applied to the base \\
% Joint torques & $\boldsymbol{\tau}$ & Teacher Only & Torques Commanded/applied at each joint \\
% Reference Targets & $\{\refpos,\refvel\}$ & Teacher Only & Reference targets described in \cref{sec:temporal_smoothing}\\
% \bottomrule
% \end{tabular}
% \end{table}
\subsection{Closed-loop Training Environment}

To precisely simulate the dynamic interactions in real-world collaborative carrying tasks, we set up a closed-loop training environment that explicitly models the interactions among the object, humanoid, and human, as illustrated in \cref{fig:training_setting}.
The environment consists of the humanoid, a supporting base body that simulates the human carrier, and the carried object to be transported, which is connected to the support body via a $6$-DoF joint.
Once the environment is initialized, the object is placed in the robot’s hand, and the hand joints are fixed in a predefined grasp pose.

During training, we randomly sample a goal command $\mathcal{G}$, with its range defined in \cref{tab:command}. We also sample a velocity $v^{\text{applied}}$ and apply it to the supporting base body at the end of the object opposite the robot-held end. The held end is predetermined for each object type (e.g., the far side of a box or the handle of a stretcher). 
We sample the magnitude of $v^{\text{applied}}$ from the range $(0, \mathcal{G})$ with added random noise and update it at twice the frequency of the goal command $\mathcal{G}$.
For angular velocity control, we set a target angular velocity and use a PD controller to apply torque to the support body. For height control, we randomly sample a target height for the support body within a predefined range and apply a PD-controlled force to adjust its height accordingly, without requiring the robot to maintain a fixed height.

It is worth noting that although the carried object and the support body are connected through a $6$-DoF joint, the inherent friction, damping, and joint limits ensure that any movement of the supporting base body directly affects the object. In this way, the dynamics of the support body are faithfully transmitted to the carried object. 

The optimization objectives are summarized in \cref{tab:reward}.

\section{Experiments}
\label{sec:experiments}

\subsection{Overview}
In this section, we conduct experiments to evaluate the effectiveness of \model. We aim to address the following key research questions via empirical analysis and discussion:

\begin{itemize}
    \item Does the residual teacher policy and the distillation training enable effective and compliant collaboration?
    %Is the implementation of the residual teacher and the distillation training scheme necessary and effective for achieving our task objectives?
    \item Is the architecture of \model\ designed in a concise and effective manner? 
    %Is the architecture of the \model\ model designed most concisely and effectively? Have we, at the network structure level, identified the most efficient solution for our targeted problem, given current knowledge and methodologies?
    \item Do the results demonstrate practical value in real-world scenarios, such as assisting humans in object transportation and reducing physical effort?
    %Do the results under the specified task conditions demonstrate practical value in real-world scenarios, such as assisting humans in object transportation and reducing physical effort?
    % \item \textbf{Q4}. Is our method capable of being effectively deployed in real-world scenarios and achieves the intended results in assisting humans with collaborative carrying tasks? Specifically, does it successfully reduce the physical burden on human operators, facilitate object transportation, and maintain the stability of the object's pose throughout the process?
\end{itemize}

% Furthermore, for our designed task scenario, we have established two distinct experimental settings, differentiated by the goal command component observed by the model during the collaboration process. They are referred to as \model \textbf{(Follower)} and \model \textbf{(Leader)}.

% For the \model \textbf{(Follower)} setting, the goal command inputs for all networks—including locomotion policy, residual teacher policy, and the final deployed student policy—are masked to zero. Under this setting, all model behaviors are determined solely by the object dynamics predicted from proprioception offsets, allowing us to thoroughly evaluate the model’s understanding of object dynamics.

% For the \model \textbf{(Leader)} setting, a randomly sampled goal command is provided as input to the policy, with its value range consistent with that used for the locomotion policy. During training, the actual velocity applied to the object is within the range of (0, $\mathcal{G}$). This setting is motivated by the observation that, in real collaborative carrying tasks performed by humans, there is typically a predefined goal (such as a known destination, obstacle avoidance, or navigation information) rather than relying solely on proprioceptive sensing. Thus, we introduce this experimental condition to evaluate the model’s task performance when the transport goal can be provided by upstream modules (e.g., visual or tactile perception systems), and to assess how well the model accomplishes the task under such prior knowledge.

\subsection{Baselines}
% To answer \textbf{Q1} and \textbf{Q2} while validating the effectiveness of our methods, we designed several baseline methods for a comparative study. These baselines are intended to address the importance of our proposed method and to provide a clear reference for the performance of it.

\begin{enumerate}
\item \textbf{Vanilla MLP}: We implement the policy as an \ac{mlp}, initialize it with the weights of the whole-body controller, and train it end-to-end with \ac{ppo}.
% We remove the residual teacher policy and the distillation step, using a single \ac{mlp} (same as the student policy) trained directly with \ac{ppo}. The policy is initialized from a locomotion policy, with the goal component in its observation masked to zero as in the \textbf{(Follower)} setting. This allows for performance evaluation in a comparable environment without teacher-student distillation.

\item \textbf{Explicit Goal Estimation}: 
We replace the whole-body control command with the predicted one, remove the residual component from the teacher policy, and distill the resulting policy into the student.
% We construct an alternative pipeline where the teacher policy directly predicts the command $\mathcal{G} = [\mathcal{G}^{loco}, \mathcal{G}^{manip}]$ for the locomotion policy in an explicit manner. 

\item \textbf{Transformer}: 
We replace the student policy’s original architecture with a Transformer.
% We replace the student policy’s \ac{mlp} with a Transformer architecture to evaluate whether an \ac{mlp} is sufficiently effective for solving the current task through reinforcement learning.

% \item \model \textbf{(Follower)}-History10: We ablate the history length to 10, restricting the model to access only the most recent 0.2 seconds of observations under the control frequency of $50Hz$. This modification facilitates the investigation of whether incorporating a longer temporal context in the observation leads to significant improvement in task performance.
\end{enumerate}

% \vspace{4pt}
% \renewcommand{\arraystretch}{1.}
% \begin{tabular}{ccc}
% \Xhline{1.5pt}
% Name & Age & Gender \\
% \midrule
% 3 & 25 & Male \\
% 2 & 22 & Female \\
% 1 & 30 & Male \\
% \Xhline{1.5pt}
% \end{tabular}
% \vspace{4pt}

\begin{figure*}[t!]
    \vspace{-35pt}
    \centering
    \begin{subfigure}[b]{0.6\linewidth}
        \includegraphics[width=\linewidth]{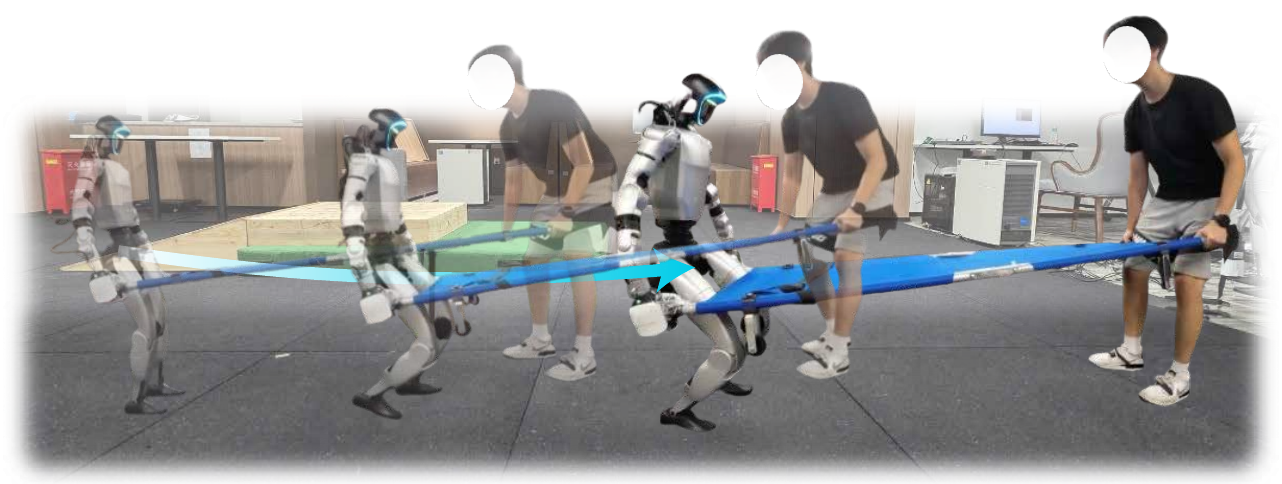}
        \caption{Stretcher Carrying}
    \end{subfigure}%
    \begin{subfigure}[b]{0.4\linewidth}
        \includegraphics[width=\linewidth]{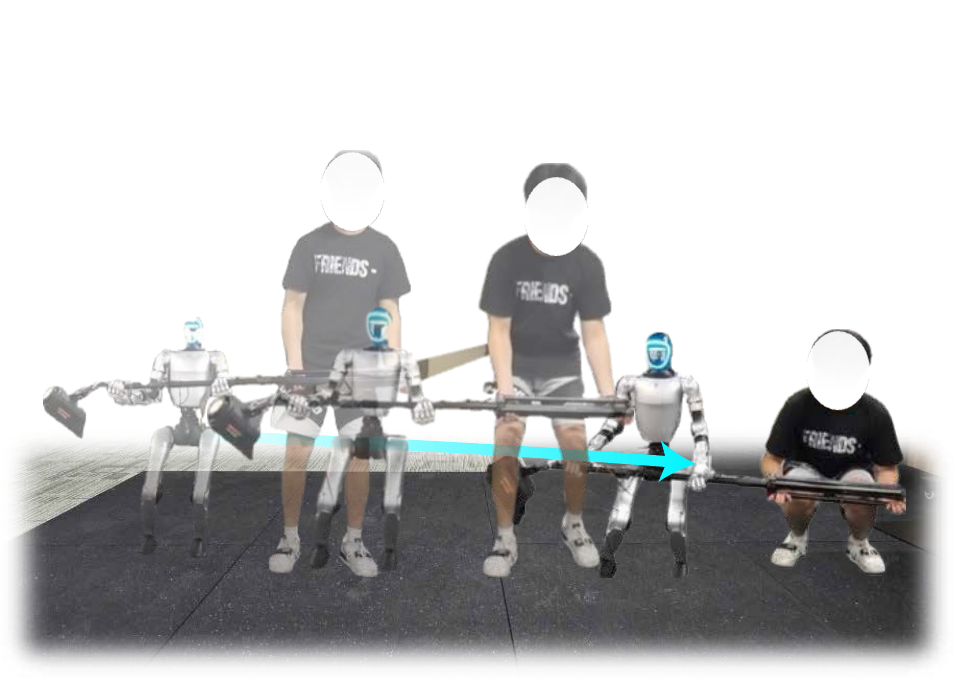}
        \caption{Rod Height Tracking}
    \end{subfigure}%
    \\%
    \vspace{-5pt}
    \begin{subfigure}[b]{0.55\linewidth}
        \includegraphics[width=\linewidth]{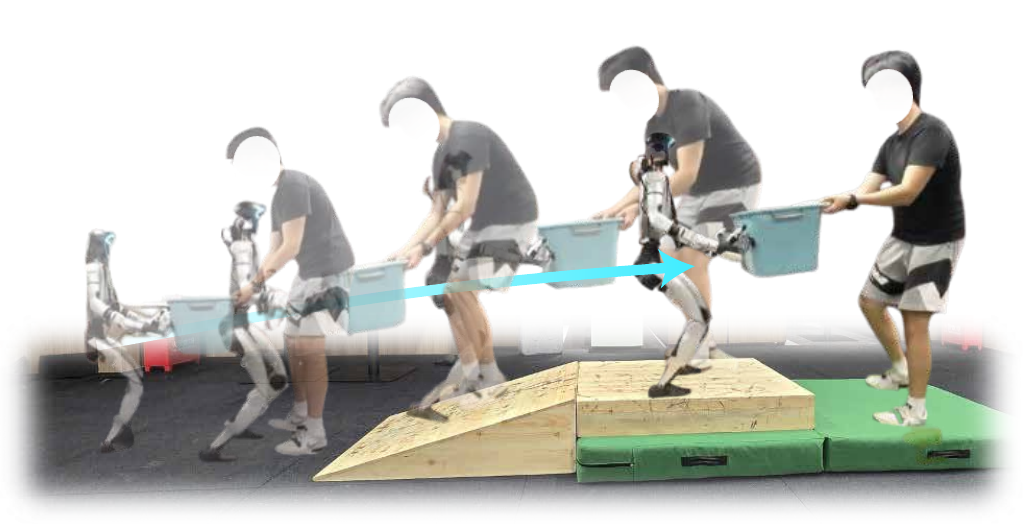}
        \caption{Box Carrying on Slope}
    \end{subfigure}%
    \begin{subfigure}[b]{0.45\linewidth}
        % \vspace{-1pt}
        \includegraphics[width=\linewidth]{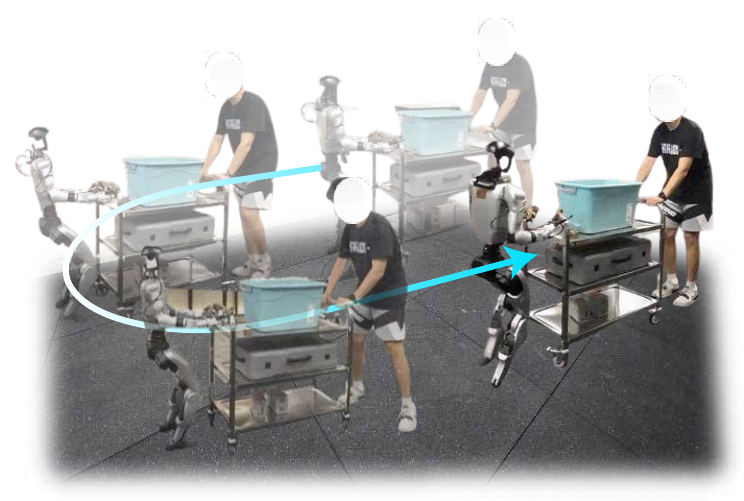}
        \caption{Cart Pushing}
    \end{subfigure}%
    \caption{\textbf{Qualitative visualizations of human-humanoid collaboration using \model.} This figure showcases our model’s ability to carry diverse objects and perform collaborative skills such as horizontal-velocity and height tracking, including under challenging conditions such as sloped terrains. The interacted objects include a $3$kg rod, an $8$kg box, an $11$kg stretcher, and a $20$kg cart, demonstrating both versatility and generalizability of the proposed method.}
    \label{fig:qual_result}
    \vspace{-15pt}
\end{figure*}

\subsection{Metrics} We use the following metrics to evaluate the performance of the proposed method in terms of trajectory following, height tracking, and coordination with the human:
\textit{Linear velocity tracking error~(Lin. Vel.)}: Mean linear velocity tracking error relative to the human over the entire episode.
\textit{Angular velocity tracking error~(Ang. Vel.)}: Mean angular velocity tracking error relative to the human over the entire episodes.
\textit{Height Error~(Height Err.)}: Height tracking error between the object ends held by the human and humanoid, measuring stability of height coordination.
\textit{Average external force~(Avg. E.F.)}: Average horizontal interaction force between the human and the object, reflecting human effort required to move the carried object along the intended direction.

\subsection{Do the residual teacher policy and the distillation
training contribute to effective and compliant
collaboration?}
We conduct simulation experiments to evaluate the performance of our model against baseline methods and analyze the choice of model architecture. As shown in the upper part of \cref{tab:main_exp_results}, our method outperforms all baselines across the evaluated metrics. The superior \textit{Lin. Vel.}, \textit{Ang. Vel.}, and \textit{Height Err.} indicate that our model achieves better collaboration with humans, while the highest \textit{Avg. E.F.} reflects the strongest compliance exhibited by the model.

Although the \textit{Vanilla MLP} achieves relatively high performance among the baseline models, it struggles to accurately track \textit{Ang. Vel.} and \textit{Height Err.}. This result indicates that linear movements are easier to infer compared to angular and vertical movements. The \textbf{teacher-student distillation framework offers a promising approach to learn these complex interaction patterns using privileged information} that is difficult to acquire directly.

While the \textit{Explicit Goal Estimation} baseline performs the poorest, this result highlights that collaborative carrying is not merely a task of predicting whole-body control commands. The dynamic interactions among the humanoid, object, and human introduce additional challenges for human-humanoid collaboration. Consequently, \textbf{implicitly learning object dynamics and human movements within a closed-loop environment proves more effective for maintaining object stability and achieving coordinated collaboration}.

\subsection{Is the architecture of \model\ designed in a concise and effective manner?}

The lower part of \cref{tab:main_exp_results} presents an ablation study on different choices of our model architecture of the student policy.
\model\ outperforms the \textit{Transformer}, which requires twice the number of training steps to converge, demonstrating that \textbf{a compact model can achieve superior performance}. This is likely because the Transformer's long-term temporal processing introduces unnecessary complexity, whereas the \ac{mlp}-based model adapts more promptly to diverse human movements, which is crucial for collaborative tasks.
For example, when the robot collaboratively carries an object with a human and they begin moving from a stationary state, the robot should focus on the current object motion rather than earlier frames that encouraged it to stay still. Relying on outdated information can cause the robot to hesitate between continuing its movement or stopping, thereby degrading the smoothness of human–humanoid cooperation.

We also ablate the history length of \model. We find that a shorter history provides insufficient information for the policy to implicitly learn collaboration with human movements from state observations. Increasing the history length to $50$ yields little improvement. We therefore select $25$ as a balance between performance and learning efficiency. This suggests that the task is not highly sensitive to long-term joint position changes, consistent with our earlier findings on the Transformer-based student policy baseline.

\begin{figure}[t!]
    \centering
    \begin{subfigure}[b]{\linewidth}
        \includegraphics[trim=3cm 9.5cm 3cm 8.0cm, clip, width=\linewidth]{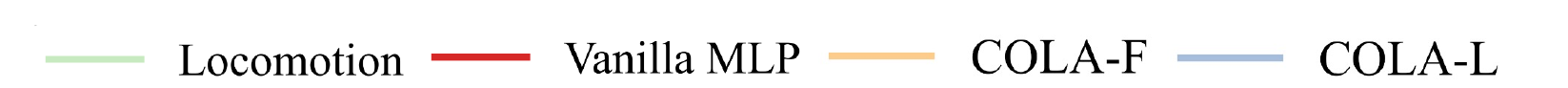}
    \end{subfigure}
    \begin{subfigure}[b]{0.5\linewidth}
        \includegraphics[trim=3cm 9.5cm 2.5cm 9cm, clip, width=\linewidth]{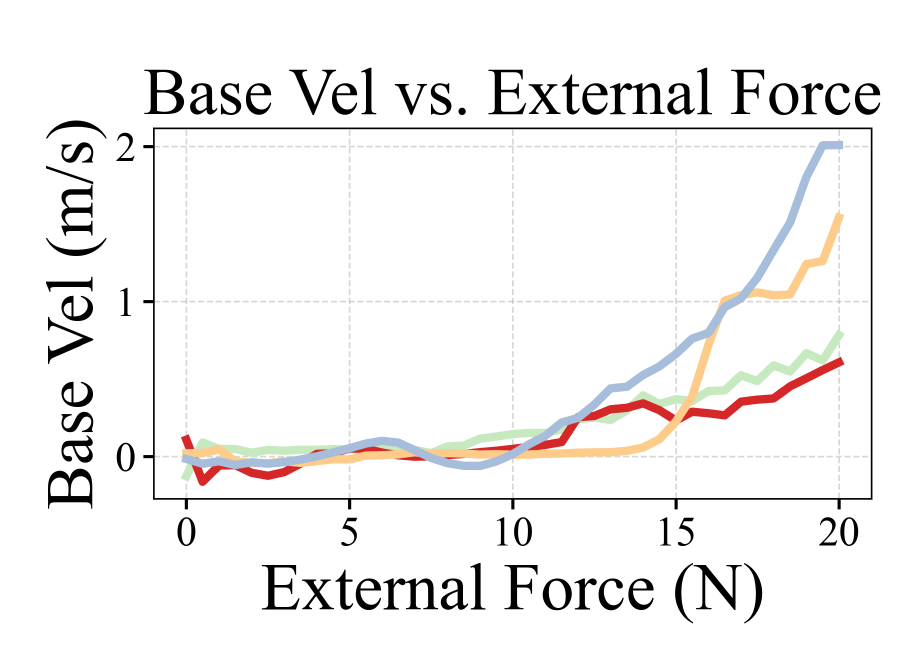}
        \caption{Sim Force}
        \label{fig:result_a_a}
    \end{subfigure}%
    \begin{subfigure}[b]{0.5\linewidth}
        \includegraphics[trim=2.5cm 9.5cm 3cm 9cm, clip, width=\linewidth]{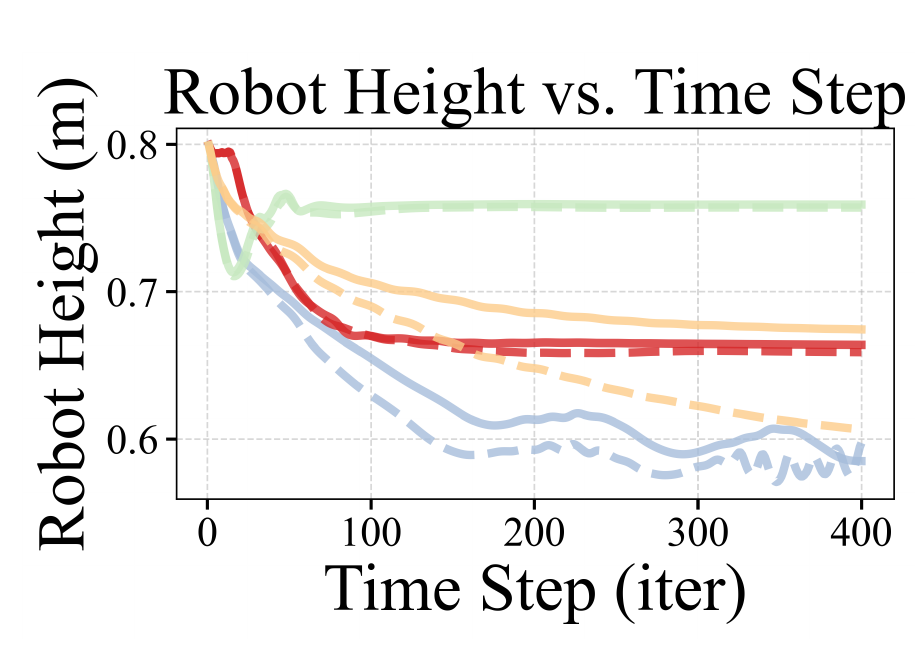}
        \caption{Sim Height}
        \label{fig:result_a_b}
    \end{subfigure}%
    \\
    \begin{subfigure}[b]{0.5\linewidth}
        \includegraphics[trim=2.5cm 9.5cm 3cm 9cm, clip, width=\linewidth]{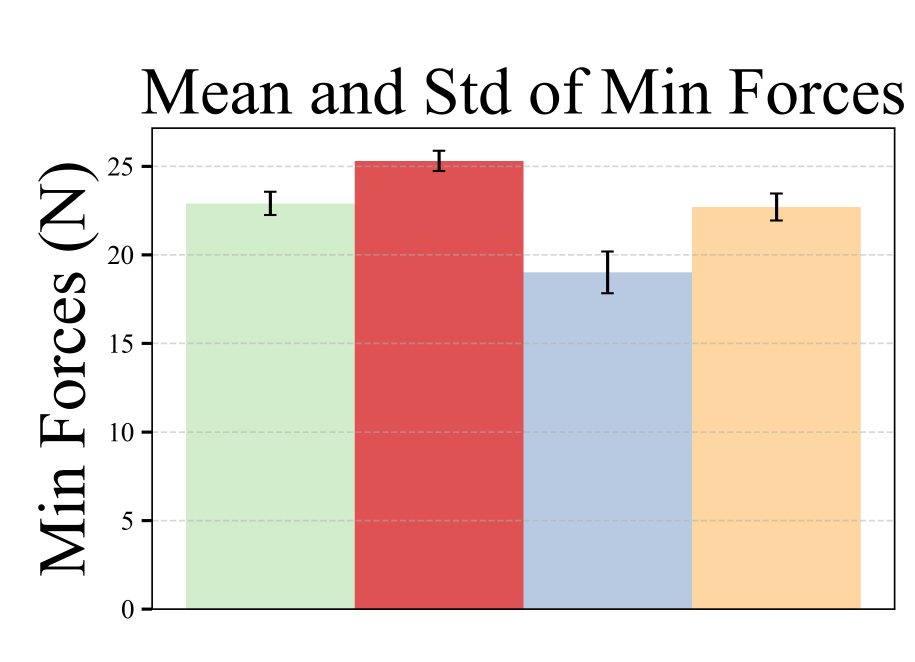}
        \caption{Real Force}
        \label{fig:result_a_c}
    \end{subfigure}%
    \begin{subfigure}[b]{0.5\linewidth}
        \includegraphics[trim=3cm 9.5cm 2.5cm 9cm, clip, width=\linewidth]
        {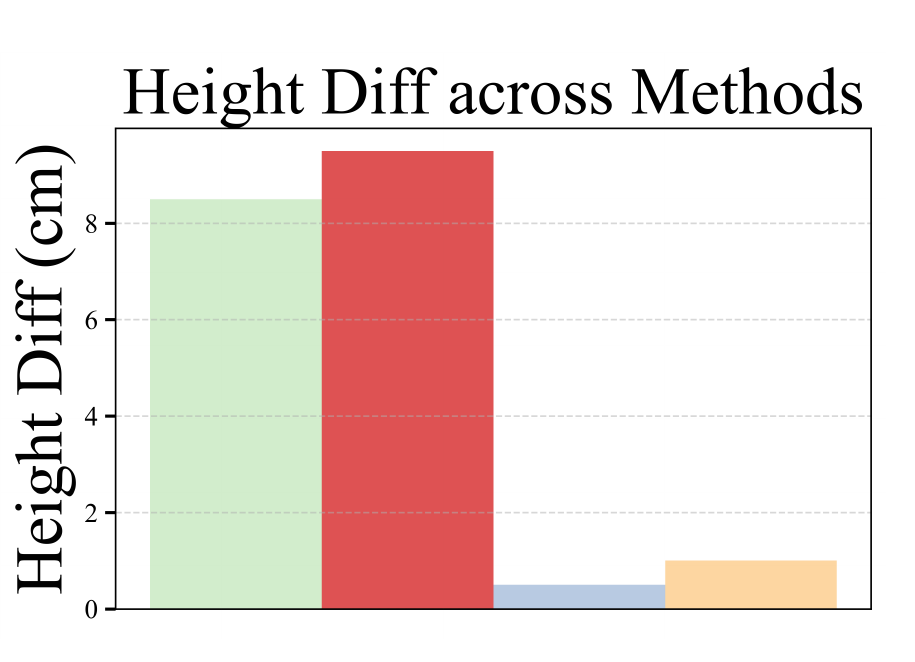}
        \caption{Real Height}
        \label{fig:result_a_d}
    \end{subfigure}%
    \caption{\textbf{Quantitative Results on the Effectiveness of Collaborative Carrying.}  (a) Illustrates the robot's velocity with a force applied to the palm of the robot where the force is linearly increased throughout a time sequence of $10.0$ seconds. (b) Illustrates the height of the robot's pelvis over time when applying external forces on the end-effector of the robot. The solid lines indicate the robot’s height in the simulator under $10N$ external forces on each palm, while dashed lines show the height under $20N$ forces. (c) Illustrates the minimal force required to move the robot in the real world. (d) Illustrates the height difference between the human-held end and the humanoid-held end of the object in real-world experiments.}
    \label{fig:result_a}
    \vspace{-12pt}
\end{figure}

\subsection{How compliant is the model to external forces in both simulation and real-world experiments?}

The results in \cref{fig:result_a_a} and \cref{fig:result_a_b} illustrate how the velocity and height respond to external forces applied along the humanoid’s x-axis. 
In \cref{fig:result_a_a}, the baseline model remains nearly stationary, whereas \model begins to follow the external force once it exceeds $15$N. Forces below $15$N are interpreted as cues for adjusting motion to stabilize the humanoid rather than for initiating movement.
\cref{fig:result_a_b} illustrates how the humanoid responds to externally applied vertical forces. The \textit{Locomotion} policy maintains a nearly constant height under the applied force, while the \textit{Vanilla MLP} squats to a fixed height regardless of the magnitude of the ascending force. This indicates that the \textit{Vanilla MLP} only supports the external force without actively complying in the vertical direction. In contrast, both \model settings effectively respond to the applied force, demonstrating agile full-body motions that comply with vertical disturbances.

It is observed that \modell consistently outperforms \modelf. We attribute this to the \textbf{goal command, which helps the policy learn to collaborate with humans more actively and precisely}. When noise and disturbances are consistently present in dynamic interactions, the goal command provides additional informative cues that enhance human-humanoid collaboration. 

\begin{table}[t!]
    \centering
    \caption{\textbf{Human study results} evaluated by 23 participants on the performance of \textit{Height Tracking} and \textit{Smoothness}.}
    \label{tab:user_study}
    \begin{tabular}{ccc}
        \toprule
        Methods & Height Tracking $\uparrow$ & Smoothness $\uparrow$\\
        \midrule
        Locomotion & $2.96$ & $2.61$ \\
        Vanilla MLP & $3.09$ & $3.09$ \\
        % lin\_vel\_z &  & $-1.0$\\
        \model & $\mathbf{3.96}$ & $\mathbf{3.96}$ \\
        % \model \textbf{(Leader)} &  & \\
        \bottomrule
    \end{tabular}
\end{table}

\subsection{Do the results demonstrate practical value in real-world scenarios?}
We also conduct real-world experiments to validate the effectiveness of our approach in practical scenarios. Qualitative results are shown in \cref{fig:qual_result}, demonstrating that our method achieves promising performance in carrying diverse objects (\eg, boxes, carts, and stretchers) across various grasping poses, even on slopes. 
Additionally, we demonstrate that our model implicitly learns to interpret human intentions through force-based interaction. When a human applies directional forces to guide the robot, the humanoid infers the desired movement command and continues executing that motion autonomously. This force-aware adaptation capability significantly broadens the range of intuitive human-robot collaboration scenarios.
These results highlight performance that surpasses existing methods.

We quantitatively evaluate our model using the following metrics:
\textit{Min. Force}: The minimum force required to move the robot, reflecting the robot’s compliance at the start of collaborative transportation.
\textit{Height Diff.} The difference in height between the two ends of the carried object, held by the human and the humanoid, reflecting the height-tracking performance.

\begin{figure}[t!]
    \centering
    \includegraphics[width=\linewidth]{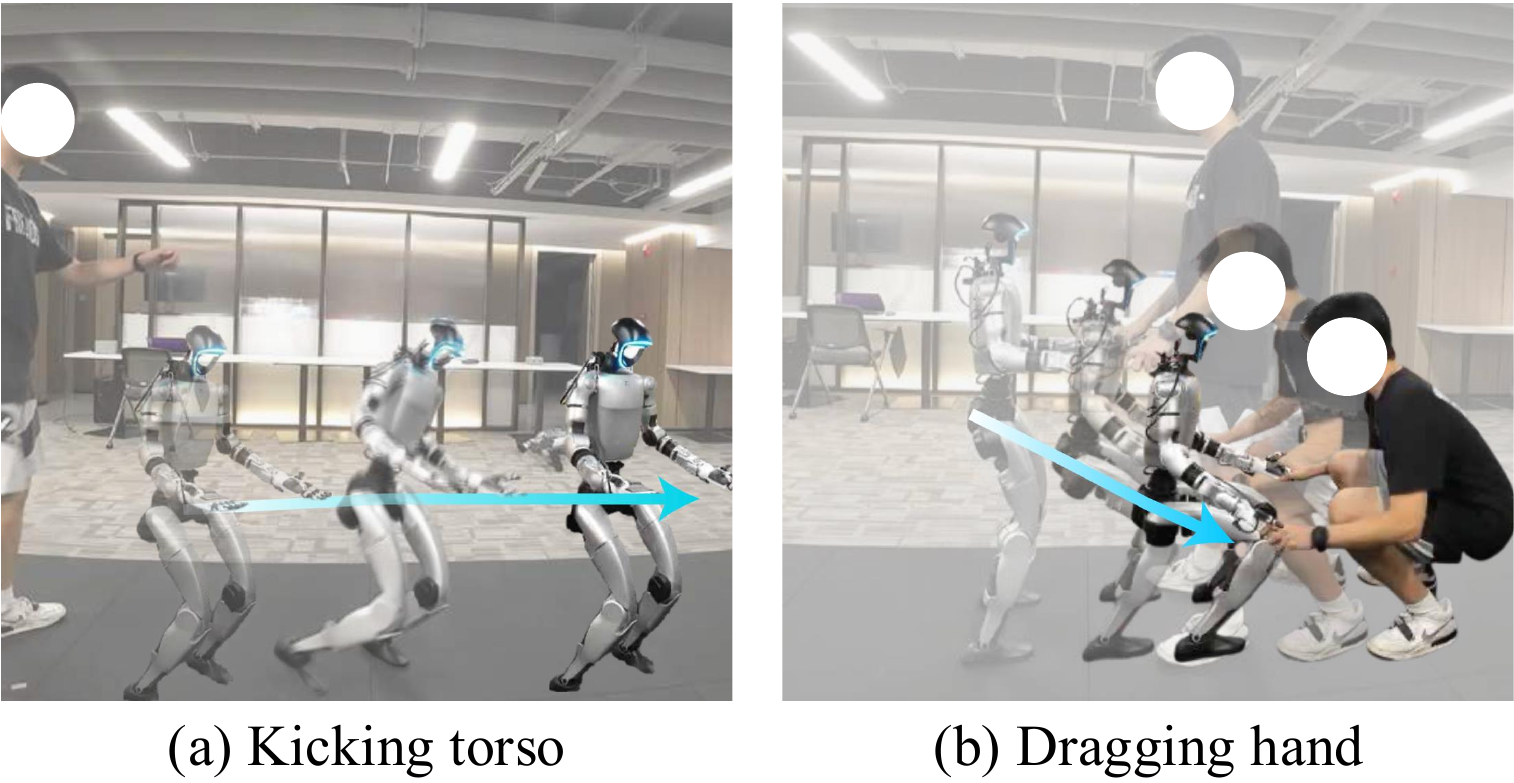}
    \caption{\textbf{Movement Analysis.} When a continuous external force is applied to the robot’s torso, it resists to maintain a stable stance. In contrast, when a smaller force is applied to the robot’s end-effector, it tends to follow the force.}
    \label{fig:sensitive}
    \vspace{-17pt}
\end{figure}

As shown in \cref{fig:result_a_c} and \cref{fig:result_a_d}, our model \textbf{reduces height-tracking error by approximately three-quarters} and demonstrates stronger compliance in following external forces compared to the baseline. These findings highlight that our model provides practical assistance for collaborative carrying tasks in real-world deployment by maintaining stable object pose tracking while reducing the physical effort required from the human operator.

We also recruited 23 participants to rate the compliance and height-tracking ability of \model during the carrying process on a scale from 1 to 5. To reduce personal bias and ensure fair evaluation, we recorded the collaboration videos, shuffled their order, and published them online for participants to provide ratings.
The results of our user study, shown in \cref{tab:user_study}, indicate that \model achieves the highest performance in both metrics. These findings further demonstrate that our model provides effective assistance to humans in real-world scenarios.

\subsection{Implicitly estimating interaction forces from joint states.}
We observe that the humanoid’s behavior during transportation is primarily sensitive to forces applied at specific joints. As shown in~\cref{fig:sensitive}, when humans apply forces to the hand or arm during carrying, the humanoid tends to follow the human’s lead. In contrast, when forces are applied to other joints, such as the torso or legs, the humanoid remains stable. These results demonstrate that our model \textbf{effectively learns the interaction dynamics among the humanoid, object, and human through the offsets between joint states and their targets}.

% In addition to requiring the robot to correctly perform collaborative carrying tasks and minimize human effort, we introduce additional qualitative evaluation methods to ensure that the robot’s understanding of the task is accurate. Specifically, we examine whether the robot’s behavior during transportation is sensitive only to specific joint positions. For example, forces applied to the hand or arm during the carrying process should be interpreted as conveying the necessary information about the object's movement, whereas forces applied to other joints, such as the torso or legs, should be considered as extraneous disturbances. We conduct experiments to assess whether the robot responds appropriately only to external forces applied to targeted joints during the carrying task. The results demonstrate that our approach successfully achieves the desired effect.

\section{Conclusions and Limitations}
In this paper, we present a unified approach for human-humanoid collaboration featuring a three-step residual learning framework that enables the humanoid to operate in two cooperation modes: leader or follower. Our method distills privileged object state information into a student policy that operates solely on proprioceptive feedback, enabling compliant and generalizable whole-body coordination without external sensors. We also propose a closed-loop training environment that explicitly models humanoid-object interactions, allowing the robot to infer human movement and adapt through compliant collaboration implicitly.

While our approach achieves effective human-humanoid collaboration through proprioception, multi-modal perception for human-humanoid collaboration remains worth exploring, as visual and tactile sensors provide more informative cues. Moreover, enabling humanoids to plan autonomously to assist humans is also valuable for future research. We hope our research provides insights for human-humanoid interaction and opens new directions for human-humanoid collaborative applications.

{
\bibliographystyle{IEEEtranS}
\bibliography{reference_header,reference}
}

\end{document}